\documentclass{bmvc2k}

\usepackage{times}
\usepackage{graphicx}
\usepackage{amssymb}
\usepackage{bm}
\usepackage{url}
\usepackage{hyperref}
\usepackage{algorithmic}
\usepackage{algorithm}
\usepackage{epsfig}
\usepackage{amsthm}
\usepackage{amsmath}
\usepackage{ducksay}
\usepackage{makecell}
\usepackage{comment}
\usepackage{paralist}
\usepackage{gensymb}
\usepackage[percent]{overpic}
\usepackage{color,soul}

\title{Vision-based system identification and 3D keypoint discovery using dynamics constraints}

\addauthor{Miguel Jaques}{m.a.m.jaques@sms.ed.ac.uk}{1}
\addauthor{Martin Asenov}{m.asenov@ed.ac.uk}{1}
\addauthor{Michael Burke}{michael.g.burke@monash.edu}{2}
\addauthor{Timothy Hospedales}{t.hospedales@ed.ac.uk}{1}

\addinstitution{
 University of Edinburgh\\
 Edinburgh, UK
}
\addinstitution{
 Monash University\\
 Melbourne, Australia
}

\runninghead{Jaques, Asenov, Burke, Hospedales}{Vision-based system identification}

\DeclareMathOperator*{\argmax}{arg\,max}
\newcommand{\name}{V-SysId}
\newcommand{\keypoint}[1]{\noindent\textbf{#1}\quad}
\newcommand{\doublecheck}[1]{\textcolor{black}{#1}}
\newcommand{\cut}[1]{}

\newcommand{\vvv}{\mathbf{v}}

\newcommand{\kk}{\mathbf{k}}
\newcommand{\pp}{\mathbf{p}}

\newcommand{\II}{\mathbf{I}}
\newcommand{\ttt}{\mathbf{t}}
\newcommand{\RR}{R}
\newcommand{\MM}{\mathbf{M}}
\newcommand{\ttheta}{\bm{\theta}}

\newcommand{\annot}[1] {\rotatebox{10}{\fontfamily{phv}\selectfont \makecell{#1}}}

\begin{document}

\maketitle

\begin{abstract}
This paper introduces V-SysId, a novel method that enables simultaneous keypoint discovery, 3D system identification, and extrinsic camera calibration from an unlabeled video taken from a static camera, using only the family of equations of motion of the object of interest as weak supervision. V-SysId takes keypoint trajectory proposals and alternates between maximum likelihood parameter estimation and extrinsic camera calibration, before applying a suitable selection criterion to identify the track of interest. This is then used to train a keypoint tracking model using supervised learning. Results on a range of settings (robotics, physics, physiology) highlight the utility of this approach.
\end{abstract}

\section{Introduction}

An understanding of the motion and physics of objects in the real world is a hallmark of the human visual system. Humans have the ability to identify objects and their properties (eg. mass, friction, elasticity) as they move and interact in the world, due to our intuitive understanding of common trajectories, object interactions, and outcomes. This ability is typically studied under the umbrella of \emph{intuitive physics} \cite{Battaglia2013SimulationUnderstanding,Ullman2014LearningScenes,Hamrick2016InferringSimulation,Baker2017RationalMentalizing}, and often considered a critical component for machines to be able to think more like humans. In the context of machine learning systems, this ability can be distilled to a requirement for \emph{unsupervised} 3D object localization and physical parameter estimation (also known as system identification) from a sensory stream, subject to some inductive bias or intuitive physics prior.

\begin{figure}[t]
    \centering
    \includegraphics[width=0.47\textwidth]{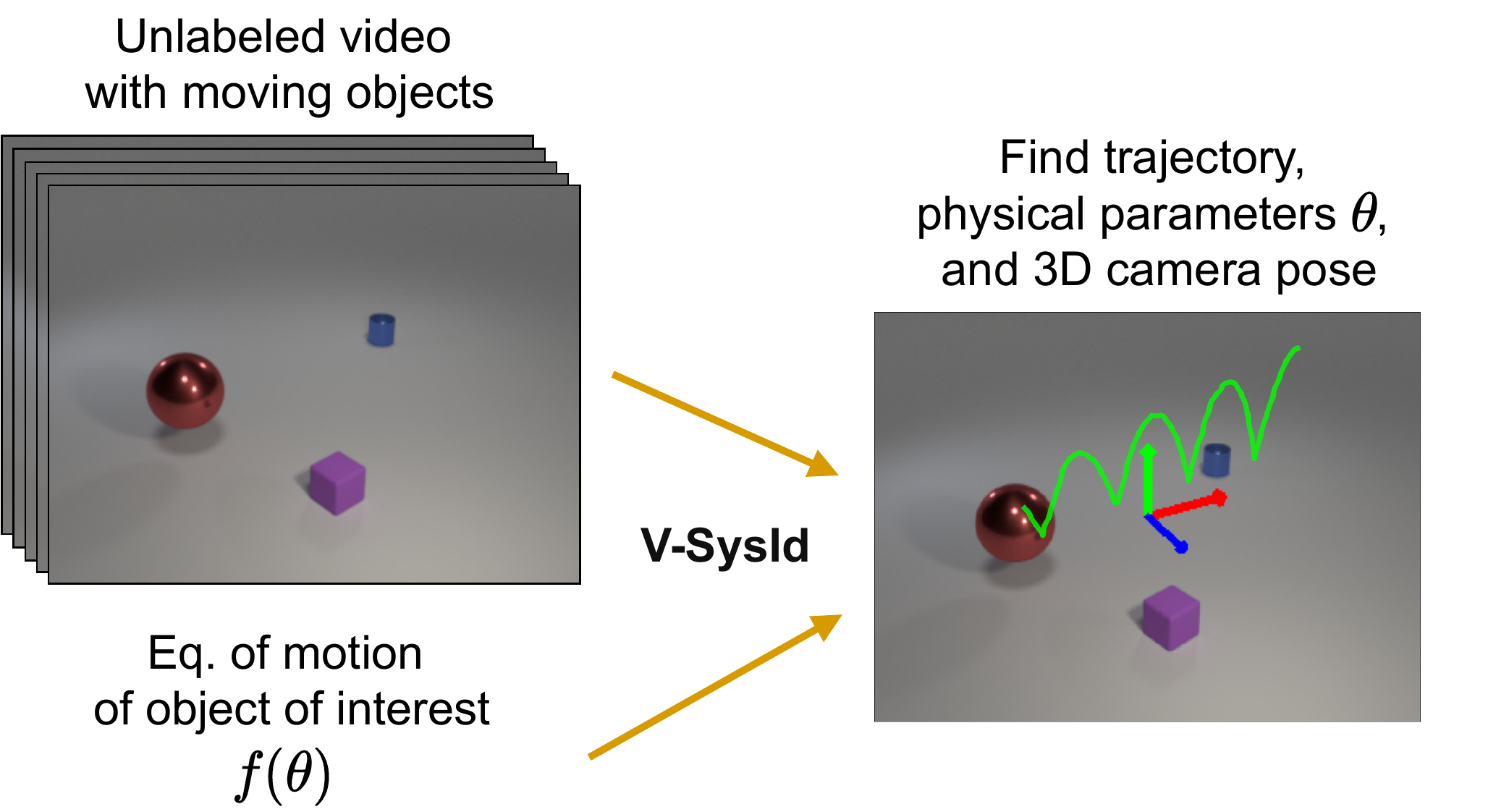}
    \caption{\cut{Problem statement.}Given an unlabeled video containing moving objects and an equation of motion, our V-SysId identifies the trajectory of the object of interest, along with its physical parameters (e.g. restitution coefficient, initial height), and 3D pose relative to the camera.}
    \label{fig:problem_diagram}
    \vspace{-3mm}
\end{figure}

Taking inspiration from this view, this paper introduces V-SysId, a novel method that enables simultaneous keypoint discovery, 3D system identification, and extrinsic camera calibration from a single unlabeled video taken from a static camera, using only the family of equations of motion of the object of interest as weak supervision. Crucially, our approach is able to identify the correct object(s) in a scene even in the presence of other moving objects or distractors. This property is key, as it greatly increases applicability to cluttered real world environments.  This allows us to perform queries such as ``\emph{find the 3D location of the bouncing ball, and determine its restitution coefficient}'' (Fig.~\ref{fig:problem_diagram}).

\begin{figure*}[!th]
    \includegraphics[width=\textwidth]{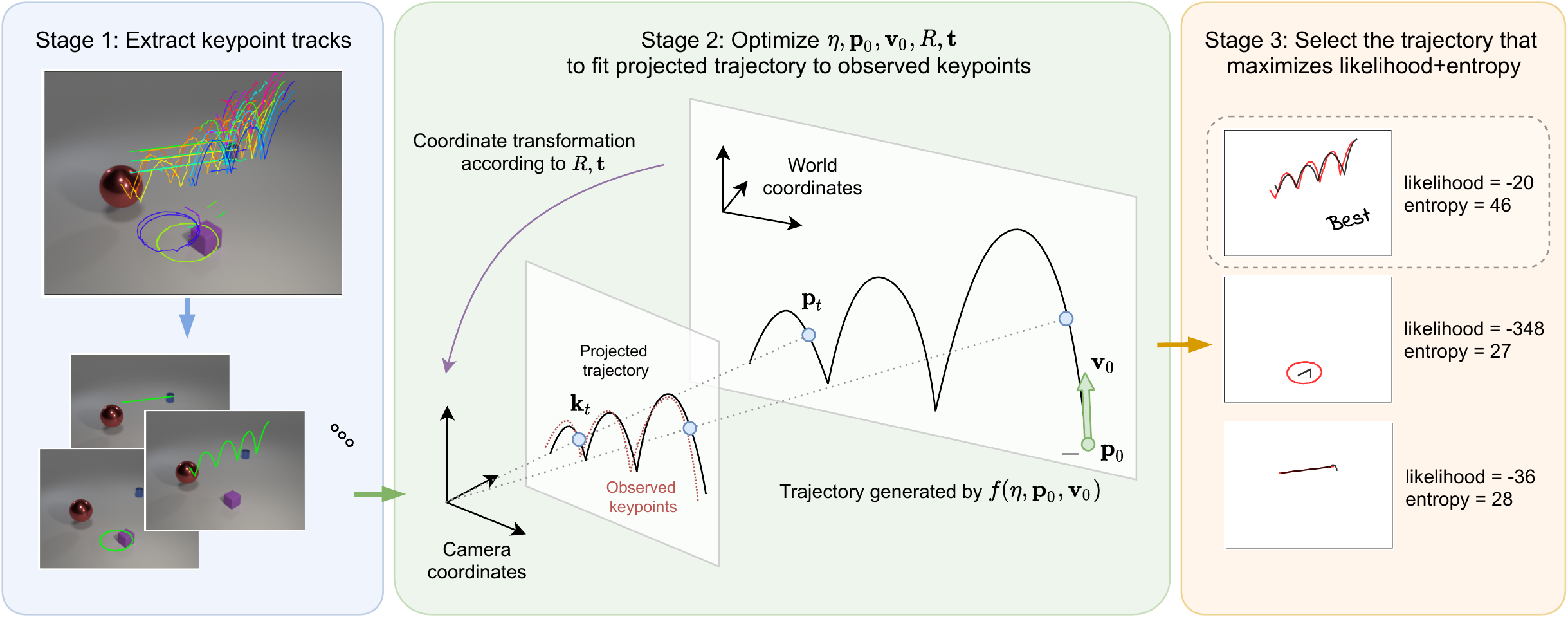}
    \vspace{-4mm}
    \caption{Our \name{} comprises 3 stages. Stage 1 extracts keypoint tracks from a video using a grid keypoint detector + KLT tracking. Each of these 2D tracks is passed to Stage 2, where the physical parameters $\theta=\{\bm{\eta}, \pp_0, \vvv_0\}$ of the 3D equation of motion $f$, and the camera pose parameters $R, \ttt$ are optimized in order to minimize the difference between the projected 3D trajectory (black, Stage 2) and the 2D keypoint track observed (red, Stage 2). Stage 3 chooses the best trajectory and corresponding parameters as those which maximize the sum of projected likelihood and a trajectory entropy criterion. Here, a bouncing ball scene with 2 moving distractors is shown, where the bouncing ball is correctly discovered as the object that corresponds to the highest entropy motion that fits the equation of motion $f$.}
    \vspace{-3mm}
    \label{fig:diagram}
\end{figure*}

V-SysId follows a 3-stage process of keypoint track proposal, optimization, and selection (Fig.\ \ref{fig:diagram}). The optimisation process alternates between maximum likelihood extrinsic camera calibration and maximum likelihood physical parameter estimation for motion tracks detected in video. This joint optimisation can be unstable, which we address through the inclusion of a curriculum-based optimisation strategy, alongside a maximum entropy criterion for keypoint identification. A key benefit of \name{} is that a neural network is \emph{not} needed for discovery or system identification in our pipeline. This means that \name{} enables keypoint discovery with high-resolution images; and can also perform system identification in \emph{single} videos, without the need to obtain large datasets, which is particularly useful in robotics applications, where data collection for neural network training can be laborious and time-consuming. The keypoints discovered by \name{} can be used as pseudo-labels to train a supervised keypoint detector, for downstream tracking or control.


These properties provide significant flexibility to \name{}, enabling its use in real world environments with important applications for control, physics understanding, and health monitoring. Specifically, we show that the \name{} can be applied to end-effector localization and extrinsic camera calibration, bouncing ball discovery and physical property estimation, and breathing frequency estimation from chest videos - all unlabeled and without regions of interest provided a priori. This is made possible by the fact that \name{} identifies keypoints belonging to objects of interest present in scenes, while ignoring any other moving objects or artifacts that do not follow the expected dynamical constraints. This alleviates the need for hand-crafted object segmentation methods or tricks to selectively remove parts of the image that may contain moving distractors; and allows keypoint discovery at a fraction of the computational expense of unsupervised neural methods that learn to identify and model every moving object in an image.

\section{Related Work}

\keypoint{System identification} and physics understanding are key to allow machine learning agents to interact with the real world. 
System identification is typically performed using proprioceptive trajectory data directly, and there has been extensive research across a range of fields  \cite{juang1985eigensystem,brincker2001modal,brunton2016discovering,Wu2015GalileoLearning,Wu2017LearningDe-animation,Li2020VisualModels} in support of this. Recent contributions include developments in physical parameter estimation \cite{Belbute-Peres2018End-to-EndControl,Cranmer2020DiscoveringBiases}, simulator learning \cite{Qiao2020ScalableControl,Sanchez-Gonzalez2020LearningNetworks}, simulation alignment for robot interaction \cite{Asenov2019Vid2Param:Video}, trajectory generation \cite{jegorova2020adversarial} and compositionality \cite{Abraham2017Model-BasedOperators,li2019learning}.

Unsupervised system identification from vision is a recent area of research that removes the requirements for trajectory data, with approaches including unsupervised physical parameter estimation \cite{Jaques2020Physics-as-Inverse-Graphics:Video,Kandukuri2020LearningPhysics,Murthy2020GradSim:Control}, structured latent space learning \cite{Karl2017DeepData,Guen2020DisentanglingPrediction,Jaques2021NewtonianVAE:Spaces}, and Hamiltonian/Lagrangian learning \cite{Greydanus2019HamiltonianNetworks,Toth2020HamiltonianNetworks,Zhong2020UnsupervisedControl}. Unfortunately, these approaches are still relatively limited in the complexity of scene they can model, and typically restricted to toy problems and simulated environments. In this work we aim to improve upon \citep{Jaques2020Physics-as-Inverse-Graphics:Video,Kandukuri2020LearningPhysics,Murthy2020GradSim:Control}'s limitation to simulated environments by performing physical parameter estimation on real dynamical scenes with distractors.

The seminal GALILEO model \cite{Wu2015GalileoLearning} demonstrated physical system identification and simulation alignment using the Physics101 dataset \cite{Wu2016PhysicsVideos}. A key shortcoming of  Galileo is that it assumes that the camera is parallel to the plane of motion, and relies on manually identified object tracks to lift the visual scenes onto object positions. In contrast \name{} is able to simultaneously estimate 3D trajectories and camera pose relative to the scene from arbitrary camera angles, greatly increasing its applicability to real world scenes. Furthermore, \name{} automatically identifies object tracks from keypoint proposals without needing human intervention, allowing us to automatically discover the objects of interest in video that are governed by the relevant equations of motion.

\keypoint{Keypoint discovery}
Keypoints are a natural representation for object parts, 
with keypoint detection and tracking one of the earliest and most studied areas of computer vision. Approaches like SIFT \cite{Lowe2004}, FAST \cite{Rosten2006MachineDetection} and ORB \cite{Rublee2011ORB:SURF} are still widely used to perform SLAM, SFM, VO\footnote{Simultaneous Localisation and Mapping, Structure-from-Motion, Visual Odometry.} and other tracking tasks (using, e.g. a KLT tracker \cite{Tomasi1991DetectionFeatures}). Given keypoint trajectories, the problem of inferring the 3D structure of a 2D trajectory using assumptions about the dynamics has been coined "trajectory triangulation" by \cite{Avidan2000TrajectorySequence,Kaminski2002GeneralTriangulation}, who assume that objects follow a straight-line or conic-section trajectory in 3D space, and that physical parameters can be uniquely identified using multiple cameras. In contrast, our method assumes only a single static monocular view. Other approaches to infer moving object structure using motion constraints include \cite{Fitzgibbon2000MultibodyObjects,Han2003MultipleCameras,David2004SoftPOSIT:Determination,Scaramuzza2009AbsoluteConstraints}.

When it comes to 2D keypoint discovery, several recent works have proposed neural network based methods that use a regularized reconstruction objective to discover objects of interest in an image \cite{Jakab2018UnsupervisedGeneration,Jakab2019LearningTranslation,Kulkarni2019UnsupervisedControl,Minderer2019UnsupervisedVideos,Gopalakrishnan2020UsupervisedPredictability,Das2020Model-BasedDemonstrations}, which can be used for downstream control tasks. However, these approaches lack the ability to estimate keypoint depth, limiting their application in realistic control scenarios. Even though these approaches obtain semantically meaningful keypoints (and in some instances are able to ignore scene objects with unpredictable motion \cite{Gopalakrishnan2020UsupervisedPredictability}), they require visual inspection in order to obtain interpretability. In contrast, \name{} provides equation-driven keypoint discovery, ensuring a known semantic meaning for learned keypoints.
A parallel stream of research tackles this from a geometric perspective, where 3D keypoints are inferred using camera motion cues or geometric constraints \cite{Suwajanakorn2018DiscoveryReasoning,Jau2020DeepConstraints,Vijayanarasimhan2017SfM-Net:Video,Wei2020DeepSFM:Adjustment}. Even though this approach has been used in complex real world settings, these keypoints lack semantic meaning, making these unsuitable for semantic discovery queries (eg. ``find the bouncing ball following these dynamics'').
%
%
%

The use of dynamics as a learning constraint has not been explored in keypoint discovery literature to date. This work proposes a method to integrate dynamical inductive biases into the keypoint discovery process, enabling extrinsic camera calibration and physics-guided discovery of objects of interest alongside the corresponding physical parameter estimation.


\section{Method}

Our goal is to discover the 3D trajectory of an object of interest in a video with possibly many moving objects, given only its family of motion dynamics, $f$. To this end, we must estimate: a) 2D keypoint locations $\kk_t$ of the object of interest in each frame $\II_t$; b) physical parameters and initial conditions $\ttheta$, of the equation of motion $f(\ttheta)$; and c) camera rotation and translation relative to the scene $[\RR, \ttt]$.
Joint \cut{end-to-end} estimation of these quantities would be 
intractable\cut{very challenging due to intractability}, so we split the objective into tractable components.
Our method, \name{}, has 3 stages 
(Fig. 1). We first describe the physical parameter+camera pose estimation stage.\cut{, as it is helpful for understanding the trajectory proposal stage.}

\subsection{Physical parameter and camera pose estimation}\label{sec:optimization}
\keypoint{Setup}  Let us assume we have a set of N 2D keypoint tracks $\mathbf{K} = \{\Tilde{\kk}^n_{1:T}\}_{n=1}^{N}$ across the video $\II_{1:T}$, and a family of 3D equations of motion $f$ with unknown physical parameters $\bm{\eta}$ and initial position and velocity $\pp_0$ and $\vvv_0$, respectively.  The equation $f$ can be rolled out over $T$ time steps using a standard integration method in order to obtain a 3D trajectory $\pp_{1:T} = f(\ttheta)$, where $\ttheta = \{\bm{\eta}, \pp_0, \vvv_0\}$.

\keypoint{Objective} Our goal is to maximize the likelihood of the observed keypoint trajectory $\Tilde{\kk}_{1:T}$ w.r.t. the physical parameters and initial conditions, $\ttheta$, and the extrinsic camera rotation and translation, $[\RR \; \ttt]$:
\begin{equation}
    \ttheta^*, \RR^*, \ttt^* = \argmax_{\ttheta, \RR, \ttt} \; p(\Tilde{\kk}_{1:T}|\ttheta, \RR, \ttt),
\end{equation}
where we factorize the trajectory likelihood as: 
\begin{equation}
    p(\Tilde{\kk}_{1:T}|\ttheta, \RR, \ttt) = \prod_t p(\Tilde{\kk}_t|\ttheta, \RR, \ttt) \nonumber = \prod_t \mathcal{N}(\Tilde{\kk}_t| \kk_t(\ttheta, \RR, \ttt), \sigma^2),
    \label{eq:likelihood}
\end{equation}
and $\kk_t(\ttheta, \RR, \ttt)$ are the 2D projection of the simulated 3D trajectory (given by $f(\theta$)):
\begin{align}
    \kk_t(\ttheta, \RR, \ttt) = [\Tilde{\pp}_{x,t}/\Tilde{\pp}_{z,t}, \Tilde{\pp}_{y,t}/\Tilde{\pp}_{z,t}] \quad;\quad
    \Tilde{\pp}_t = \MM \, [\RR \; \ttt] \, \pp_t
    \label{eq:proj2d}
\end{align}
with $\MM$ being the intrinsic camera matrix. In this work we assume known camera intrinsics.

In order to reduce the space of possible solutions (and therefore local minima) of Step 1 above, we restrict the camera rotation matrix $\RR$ to have $\text{roll}=0$. This means the camera cannot rotate about its projection axis, which is the case in the vast majority of settings. Using the projection plane in camera coordinates as $xy$ and the projection axis as $z$, we parametrize $\RR$ as $\RR(\alpha, \beta) = \text{EulerRotationMatrix}(\alpha, \beta, 0)$, where $\alpha$, $\beta$ and $\gamma=0$ correspond to the pitch, yaw and roll Euler angles, respectively.
We found that this parametrization greatly improves results and optimization stability.

\keypoint{Optimization} To maximize \eqref{eq:likelihood} we apply an iterative optimization procedure. Given an initial estimate for $\ttheta$, $\RR$ and $\ttt$, we alternate the following steps until convergence:
\begin{compactenum}
    \item Keeping $\ttheta$ fixed, maximize \eqref{eq:likelihood} w.r.t. $\RR$ and $\ttt$ using gradient descent;
    \item Keeping $\RR$ and $\ttt$ fixed, maximize \eqref{eq:likelihood} wrt $\ttheta$ using gradient descent (with numerical or analytical \cite{Belbute-Peres2018End-to-EndControl} differentiation) or global optimizer (e.g. CEM \cite{Rubinstein1997OptimizationEvents}; BO \cite{Mockus1989BayesianOptimization}).
\end{compactenum}

Estimation of the physical parameters over the full sequence (possibly hundreds of timesteps) is prone to local minima, as the dependency on the parameters can be highly non-linear\footnote{Global optimizers have a slight advantage in this case, although they require very many iterations to find a good minimum.}. This is further affected by the use of a non-optimized camera pose at the first iteration. In order to address this, we start by performing a step of physical parameter and pose estimation on a small initial trajectory interval, $T_0$, adding $m$ points to the trajectory at each iteration, as described in the appendix, Algorithm 1.

\subsection{Trajectory proposal and Selection}\label{sec:trjProp}
\keypoint{Proposal}
In an unlabeled video, ground-truth 2D keypoints are not available, but keypoint trajectories are required to maximize the likelihood in \eqref{eq:likelihood}. Joint estimation \doublecheck{of physical parameters} with a neural network-based keypoint detector would be hard to optimize due to the difficulty of backpropagating through physics rollouts and camera projection into a CNN in a stable manner (\cite{Jaques2020Physics-as-Inverse-Graphics:Video}). Therefore, we propose a simpler, more robust approach: 
We extract keypoints from the first frame of the video using a keypoint detector, and track them using an optical-flow-based tracker. This produces a set of 2D keypoint tracks $\Tilde{\kk}_{1:T}$, and allows physical parameter+pose estimation to be performed for each track independently. \cut{Interestingly, we found that simply extracting keypoints from a uniform grid of 10x10 pixels across the image yielded more reliable tracks than using classic (SIFT, ORB, FAST), or modern (SuperPoint, LF-Net) keypoint detectors, with a fraction of the computation.} 

%
\keypoint{Selection}
Once the physical parameters and pose are estimated for each keypoint track, the best tracklet can be identified by isolating the highest projection likelihood \eqref{eq:likelihood}. However, in order to prevent trivial keypoint tracklets from being chosen (since a static keypoint will easily attain maximal likelihood), we add a temporal entropy term to the likelihood, such as the temporal standard deviation of the observed trajectory, resulting in the following selection criterion:
\begin{align}
    n^{\text{best}} &= 
    \argmax_{n\in1..N} \; p(\Tilde{\kk}^n_{1:T}|\ttheta, \RR, \ttt) + \text{Stddev}_t(\Tilde{\kk}^n_{1:T})
    \label{eq:criterion}
\end{align}
This finds the highest entropy trajectory that satisfies the physical motion constraints. 


The full V-SysId procedure is depicted in Fig.\ \ref{fig:diagram} and pseudocode is shown in Algorithm 1 in Appendix \ref{sec:pseudocode}.

\begin{figure*}[t]
    \centering
    Bouncing ball with unknown velocity, initial height, and restitution coefficient.
    \includegraphics[width=0.99\textwidth]{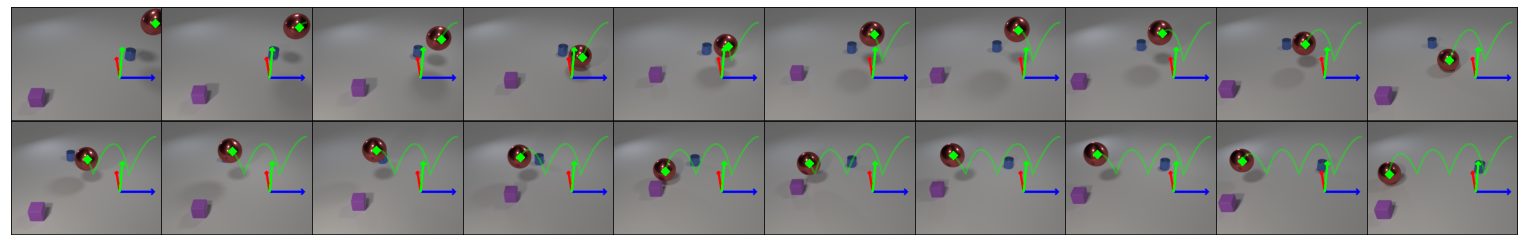}

    Archimedes spiral with unknown radius, radius increase rate, and angular velocity.
    \includegraphics[width=0.99\textwidth]{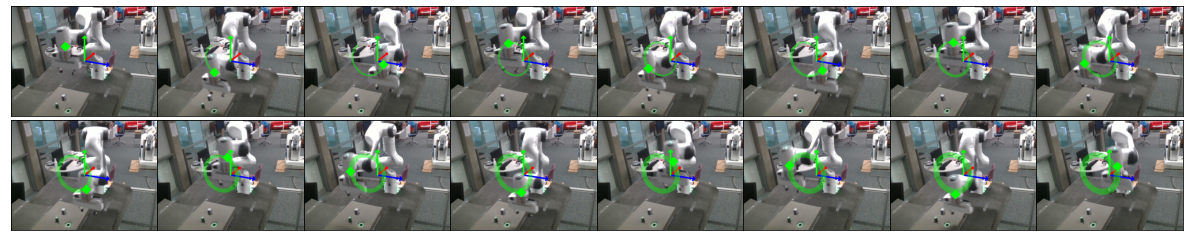}
    \caption{Discovered object and 3D perspective given the only the family of equations above as weak supervision. \textbf{Top:} Example bouncing ball scene. More scenes can be found in Fig.\ \ref{fig:bb_rollouts_appendix} in the Appendix. \textbf{Bottom:} Spiral robot arm end-effector in a real lab setting.}
    \label{fig:rollouts}
    \vspace{-3mm}
\end{figure*}

\keypoint{Inference at run-time}
Once the \name{} procedure is complete, keypoints are available for the objects of interest in each frame in the video. These can be treated as pseudo-ground truth keypoints, and used to train a neural network (or another visual object detector) by supervised learning, in order to perform fast keypoint detection at test-time. 

\section{Experiments}


\keypoint{Keypoint detection and tracking:} 
We detect keypoints in the first frame by using taking the locations of a 10x10 grid across the frame, and use the KLT algorithm to track these across the video.
 We show comparisons between grid, ORB, SuperPoint and LF-Net keypoint detectors in Appendix \ref{sec:keypoint_ablation_appendix}.

\keypoint{Track filtering:} Since the grid keypoint detector extracts hundreds of keypoints, we remove tracks whose length is less than 60\% of the full video, and whose temporal stddev \eqref{eq:criterion} is less than 10 pixel, prior to optimization. This reduces computation, as physical parameter + pose estimation is performed on only the most feasible tracks.

\keypoint{Physical parameter estimation:} The gradient-based BFGS \cite{Fletcher2000PracticalOptimization} 
is used with numerical derivatives for physical parameter optimization. Although \cite{Belbute-Peres2018End-to-EndControl} provides an elegant method for analytical differentiation through contacts, we found it much harder to implement, and ultimately slower, than simple BFGS. \cut{Additionally, \cite{Belbute-Peres2018End-to-EndControl} has only been applied to noiseless simulated environments and assumes knowledge of the true initial position and velocity of the objects, which is not available here.} Since the equations of motion considered here are planar, the $z$ component of $\vvv_0$ is constrained to $0$. The remaining parameters are learnable.

On the first iteration, the initial position $\pp_0$ is set to be the reprojection of the first 2D keypoint $\Tilde{\kk_0}$ onto the $z=5$ plane in world coordinates. This results in an initial position whose camera projection is the first keypoint. The initial velocity is $\vvv_0 = [0, 0, 0]$. We found these settings essential to avoid local minima in the incremental optimization.

\keypoint{Camera pose estimation:} BFGS is also used with finite differencing for the camera pose optimization step. The parametrization of $\RR$ on pitch and yaw provides a smooth objective that is easy to optimize, whereas we found the PnP algorithm to result in large and not necessarily optimal jumps between steps. We initialize the camera pose parameters as $\alpha=0$, $\beta=0$, and $\ttt=[0,0,0]$.

\keypoint{Curriculum-based optimization:} We use 25 input frames to start the optimization, adding 10 frames per iteration until reaching the full length of the sequence.


\begin{figure*}[th]
  \centering
    \begin{minipage}{0.49\textwidth}
        \centering
        \small{Object of interest: bouncing ball}
        \\[1mm]
        \includegraphics[width=0.6\textwidth]{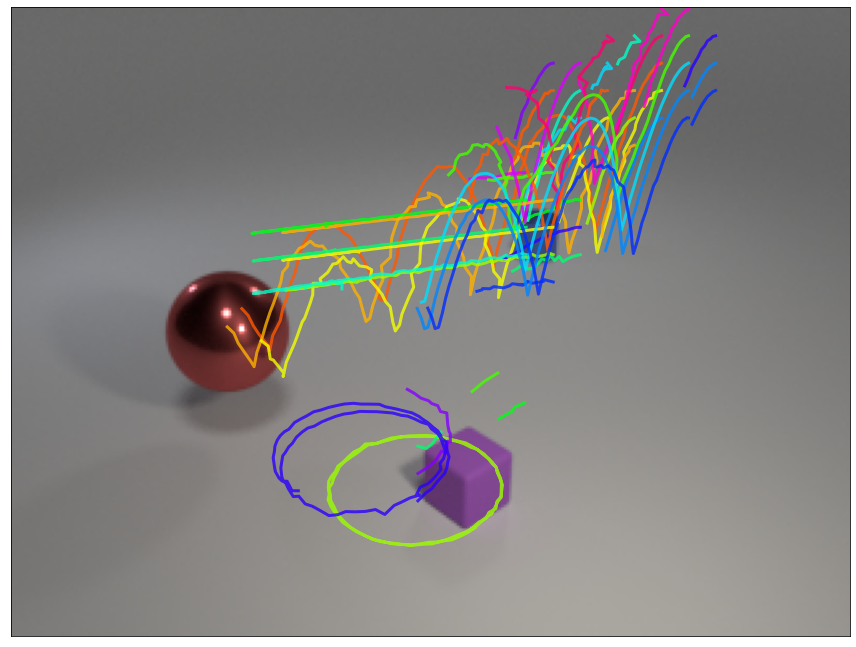}
        \includegraphics[width=0.26\textwidth]{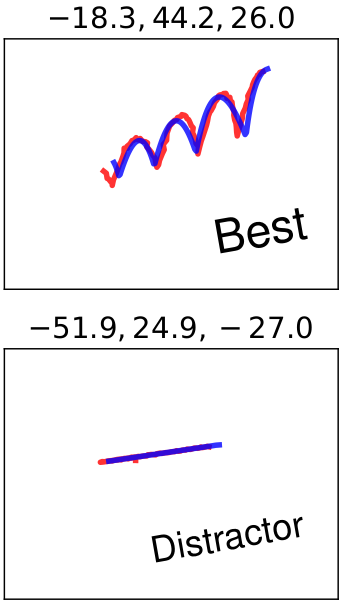}
    \end{minipage}
    \begin{minipage}{0.49\textwidth}
        \centering
        \small{Object of interest: robot arm end-effector}
        \\[1mm]
        \includegraphics[width=0.6\textwidth]{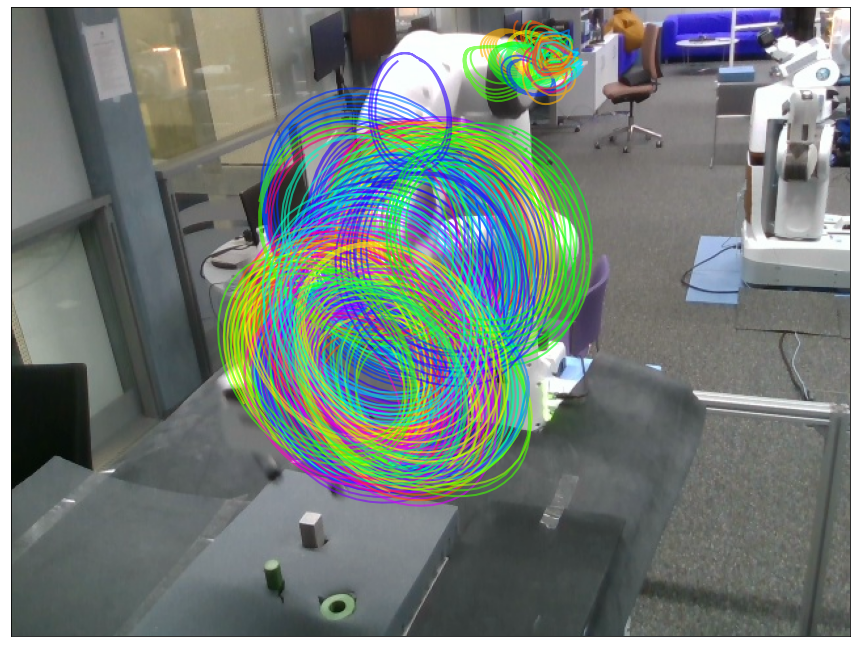}
        \includegraphics[width=0.26\textwidth]{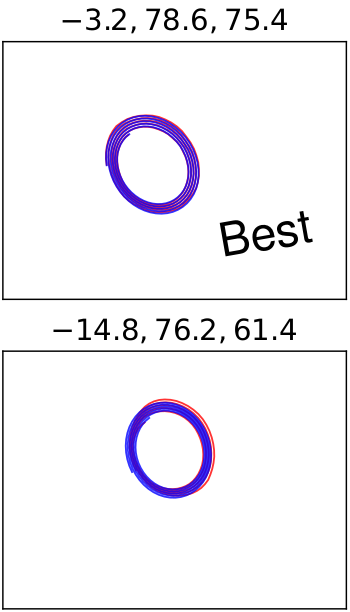}
    \end{minipage}%
    \hspace{2mm}
    \caption{\textbf{Left:} Keypoint tracks propsed by a grid keypoint detector + KLT tracker (short or static tracks not shown here for clarity). \textbf{Right:} Subset of the extracted keypoint tracks (red) and projected fitted trajectories (blue), with the corresponding projection loglikelihood, entropy, and their sum, over each plot. 
    See Fig.~\ref{fig:more_trajs_and_stats} for a more comprehensive illustration.}
    \label{fig:trajs_and_stats}
    \vspace{-4mm}
\end{figure*}

\subsection{Environments} 
\keypoint{Franka Emika Panda Robot:} This sequence consists of a multi-joint robot arm (Franka Emika Panda) in a laboratory setting, where the goal is to find the end-effector's 3D location and the camera pose relative to this. The end-effector was programmed to follow an archimedes spiral in an unknown 2D plane. The spiral is described by:
$r = a + b\cdot t \quad ; \quad \theta = \theta_0 + \omega \cdot t$
where $r$, $a$, $b$, $\theta_0$, $\omega$ are unknown parameters, to be learned by V-SysId, and $t$ is the time in seconds. A sequence of frames for this environment can be seen on Fig. \ref{fig:rollouts}, bottom. The video is 250 frames long, with a resolution of $640 \times 480$. 

\keypoint{Simulated bouncing ball:} This environment consists of a simulated bouncing ball with moving distractor objects. The bouncing ball follows the equation of motion:
\begin{align}
\vspace{-2mm}
\begin{cases}
a_y = -g \;, \;\text{if} \; y>\text{floor} \\
v_x = v_{x_0};\quad v_z = 0;\quad v_y = -\epsilon \,v_y  \;, \;\text{if} \; y=\text{floor} \\
\end{cases}
\vspace{-2mm}
\end{align}
where $a$ is the acceleration, $v$ is the velocity, $y$ is the ball height, $\epsilon \in [0, 1]$ is the restitution coefficient, and $g=9.8$ is the gravity. The ball moves in the $z=z_0$ plane with constant horizontal velocity, with the pose parameters $R, \ttt$ being responsible for correctly inferring the location of this plane relative to the camera. Photorealistic scenes are rendered in Blender following the Clevr protocol \cite{johnson2017clevr}, and trajectories are rolled out using Euler integration.

There are two distractor objects on the floor scene, one moving in a circle, and another in a straight line. This environment is used to obtain thorough quantitative results regarding the physical parameter and camera pose estimation abilities of V-SysId. To this end we generate 108 sequences along the following factors of variation: initial height; initial horizontal velocity; restitution coefficient; camera location; moving/static distractor objects. The physical parameters $y_0$, $v_{y_0}$, $v_{x_0}$, $\eta$, and floor height are unknown, and discovered by the optimization process of \name{}. The sequences are 120 frames long, with a resolution of $320 \times 240$. 

\begin{figure*}[th]
    \includegraphics[width=0.99\textwidth]{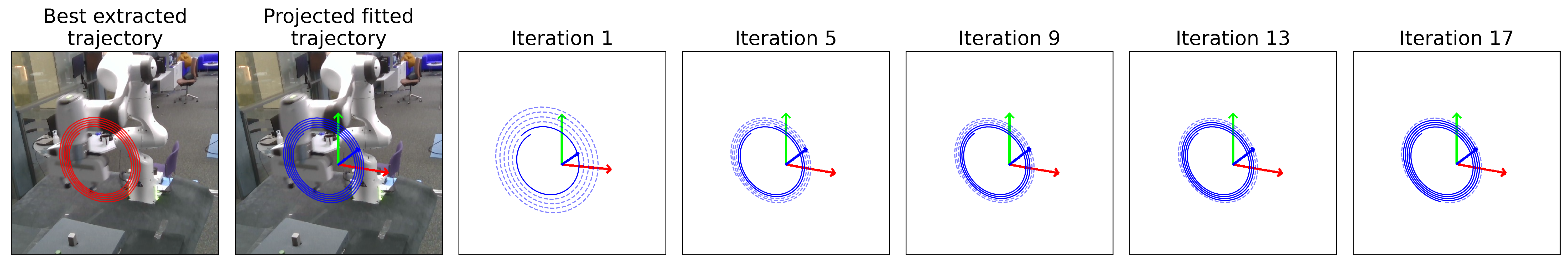}
    \includegraphics[width=0.99\textwidth]{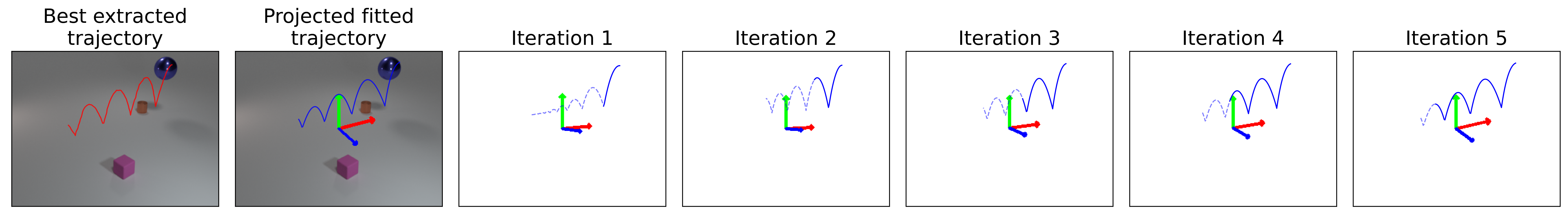}
    \vspace{-2mm}
    \caption{Visualization of the curriculum-based optimization iterations for the spiral robot (top) and bouncing ball (bottom) scenes. The red line corresponds to the extracted keypoint track and the solid blue line corresponds to the trajectory with parameters estimated so far. The dashed blue line corresponds to the predicted trajectory over the full length of the sequence, under the parameters estimated so far. We can see that the curriculum-based optimization progressively improves the physical parameter and pose estimates.}
    \label{fig:iterations}
    \vspace{-4mm}
\end{figure*}


\subsection{Visualizing keypoint proposal and optimization}

We start by visually exploring the results obtained by \name{} on the spiral robot and bouncing ball datasets. Fig.~\ref{fig:rollouts} shows the keypoints discovered for two of the scenes. These show that \name{} correctly identifies objects of interest according to the given equation of motion. 

The keypoint proposal and selection process is visualized further in Fig.~\ref{fig:trajs_and_stats}. Fig.~\ref{fig:trajs_and_stats} (left) shows the proposed keypoint tracks extracted at the proposal stage (Sec~\ref{sec:trjProp}), and Fig.\ \ref{fig:trajs_and_stats} (right) shows the results obtained by the optimization process (Sec~\ref{sec:optimization}) on a subset of these, ordered by their selection criterion score (the third number above each plot). The trajectory chosen by V-SysId according to the maximum entropy criterion 
is labeled as ``Best''. These figures highlight several important points: Firstly, \name{} is successful despite the large number of distractor keypoints from the various moving parts of the scene (most notable in the robot arm sequence). Secondly and crucially, the optimization process and the maximum entropy criterion are able to fit and identify the best trajectory, correctly discovering the object corresponding to the motion of interest. 

In order to further understand the curriculum-based optimization process, we visualize the optimization iterations of two keypoint tracks selected by V-SysId in Fig.~\ref{fig:iterations}. We can see that upon completion (2nd column), the orientation of the trajectory in 3D space is correctly identified by the model, and that each iteration progressively adjusts both the trajectory's shape (parametrized by the physical parameters) and the camera pose. This leads to a stable optimization procedure where both physical parameters and camera pose are identified.


\subsection{Evaluating parameter estimation}

Even though the scale is generally unidentifiable (this and other limitations are discussed in Sec.~\ref{sec:limitations}), in the case of a bouncing ball both the initial height and the restitution coefficient are exactly identifiable. This allows us to compare their learned values to the ground truth values used for the simulations. In addition, we can compare the camera angles identified to those used in simulation in order to evaluate the quality of the extrinsic camera calibration.

The percentage error in restitution coefficient, initial height (distance to floor), and camera angle relative to the simulation ground-truth can be seen in Table~\ref{table:params}. We can see that all parameters are found with a good degree of accuracy, with physical parameters being slightly more accurate than the camera pose. Notably, the errors are similar with and without moving distractors (within 95\% confidence intervals), showing that \name{} is able to correctly identify the object of interest even in the presence of distractor objects.

In order to highlight the importance of the curriculum-based optimization strategy, we compare the projection likelihood using our incremental alternate optimization with alternate optimization using the full sequence at every step. Averaging over the bouncing ball scenes, we obtain projection RMSE (pixels) of $-9.31$ and $-109.35$, respectively. A similar decrease in performance was observed when using CEM and BO optimizers. This shows that gradually increasing sequence length and using a gradient-based optimizer is key to the convergence of \name{}.

\begin{table}
\centering
\begin{tabular}{l| c c c }
   Distractors& \makecell{Restitution\\coefficient (\%)}& \makecell{Initial height\\in 3D (\%)} & \makecell{Camera\\angle (\degree)} \\ \hline  
 With & $3.8 \pm 1.5$ & $9.7 \pm 4.0$ & $8.0 \pm 1.8$ \\  
 Without & $2.7 \pm 0.8$ & $6.7 \pm 3.0$ &  $9.9 \pm 2.6$  
\end{tabular}
\caption{Relative error (percentage) between the ground-truth simulation physical parameters and camera pose, and those estimated by V-SysId, for the bouncing ball scene. Error bounds correspond to a 95\% confidence interval.\label{table:params}}
\vspace{-5mm}
\end{table}

\vspace{-2mm}
\subsection{Tracking by supervised keypoint detection}

Once detected, the keypoints discovered by \name{} can be used as pseudo-ground-truth to train a supervised keypoint detector. For the bouncing ball dataset, the training set consists of 2838 pseudo-labeled frames, and the test set consists of 948 hand-labeled frames from unseen scene configurations. For the robot dataset, the training set consists of 250 pseudo-labeled frames, and the test set consists of 150 hand-labeled frames from unseen end-effector positions. For the supervised keypoint detector, we use a fully convolutional neural network with 6 ReLU layers with 32 channels, with stride 2 on the 3rd layer, and 2 output channels with 2D softmax activation. These maps are converted to $[x,y]$ coordinates by taking the softmax-weighted mean over the output coordinate grid, as per \cite{Jakab2018UnsupervisedGeneration}. The input images have a downsampling factor of 4 relative to the original frame resolutions, but we report the keypoint error in the original image space. We train the networks for 20 epochs with batch size 16, and Adam \cite{Kingma} (learning rate $3\times 10^{-4}$). 

Results are shown in Table~\ref{table:supervised}. The supervised keypoint detector produces highly accurate detections, confirming the quality and usability of the keypoints discovered by \name{} even on small datasets of high-resolution scenes.

\begin{table}[!h]
\centering
\begin{tabular}{l| c }
   Environment & RMSE (pixel distance) \\ \hline 
 Simulated bouncing ball \small{($240\times320$)} & $8.41 \pm 1.50$ \\ 
 Spiral robot \small{($480\times640$)}  & $3.89 \pm 0.45$
\end{tabular}
\caption{Detection error on the held-out test set of the keypoints extracted by the inference neural network, after training using the keypoints discovered by \name{} as supervision. Bounds correspond to 95\% confidence interval.}
\label{table:supervised}
\vspace{-1mm}
\end{table}

We recommend reading of the supplementary materials for additional ablations experiments, and an application of \name{} to breathing rate estimation from video.

\subsection{ROI discovery in breathing videos using RANSAC} 
\label{sec:breathing_subsection}
\keypoint{Setup} To further demonstrate the applicability of \name{} to real world scenarios, we collected 8 videos of people breathing under different pose, lightning, clothing and distractor settings, with the goal of discovering the relevant region region of the image and using it for breathing rate identification. The true breathing rate was obtained by manual annotation. Videos contain between 150 and 300 frames, at 30 fps and 480x640 resolution. 

Unlike seminal work in video-based physiology and plethysmography \cite{boer2010slp}, \name{} does \textit{not} require careful hand selection of the regions of interest and is robust to the existence of distractor motions in the scene. \name{} simultaneously identifies the region of interest (here, the set of relevant keypoints, rather than a single one) corresponding to sinusoidal motion, and the underlying breathing rate. 

\keypoint{Results} 
We have seen how single keypoint discovery can be achieved using \name{}, but the algorithm can be easily modified to allow discovery of sets of keypoints constituting a region-of-interest. We use the chest video dataset as a prototypical application. The goal is to discover the keypoints in the video corresponding to sinusoidal motion. We start by extracting keypoint tracks as in Stage 1 of \name{} (filtering out any tracks with a temporal stddev less than 0.7), and transform these 2D tracks into 1D timeseries by taking the projection onto the 1st PCA component of the timeseries (i.e. the 2D direction of highest variance). Each timeseries is standardised, and fit to a sinusoid as per Stage 2 of \name{} (without the 3D component). In order to identify the best set of tracks, we use a RANSAC inlier count, by measuring the error between a track's fitted sinusoid and all the other extracted tracks, and considering a track an inlier if the MSE is below 0.75. The best track is chosen according to a modified maximum entropy criterion in Stage 3, where the likelihood term is replaced by the inlier count. The ROI is defined as the set of inlier tracks of the best track. 

\begin{figure}[th]
    \centering
    \includegraphics[width=0.11\linewidth]{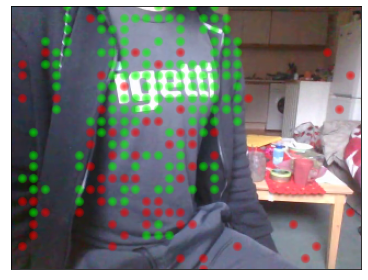}
    \includegraphics[width=0.11\linewidth]{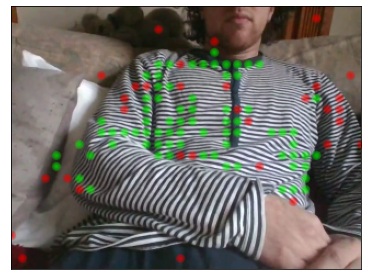}
    \includegraphics[width=0.11\linewidth]{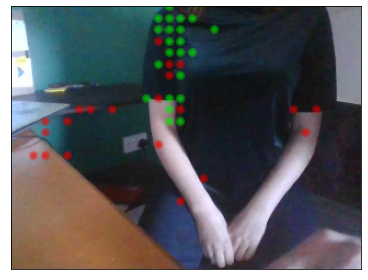}
    \includegraphics[width=0.11\linewidth]{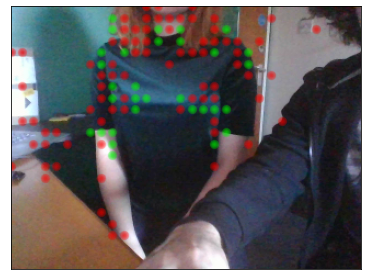}
    \includegraphics[width=0.11\linewidth]{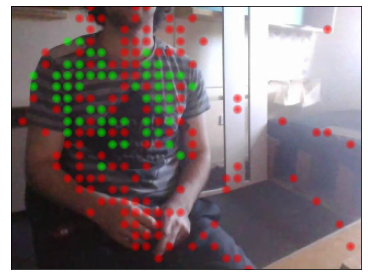}
    \includegraphics[width=0.11\linewidth]{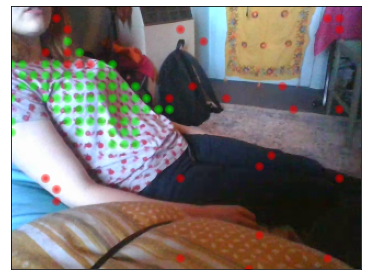}
    \includegraphics[width=0.11\linewidth]{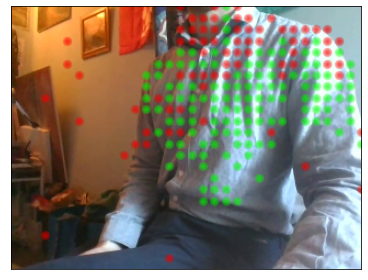}
    \includegraphics[width=0.11\linewidth]{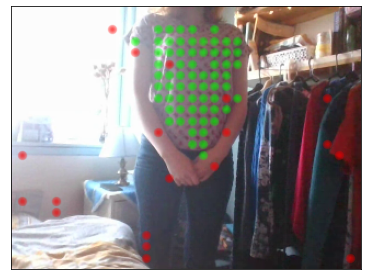}
    \includegraphics[width=0.11\linewidth]{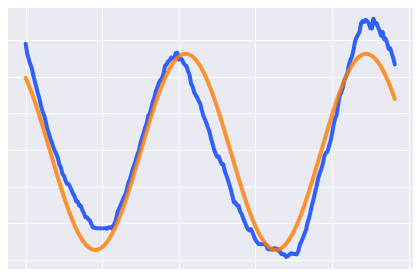}
    \includegraphics[width=0.11\linewidth]{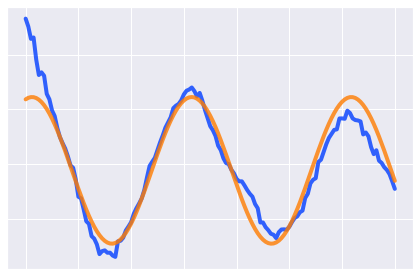}
    \includegraphics[width=0.11\linewidth]{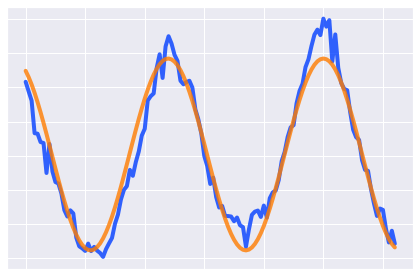}
    \includegraphics[width=0.11\linewidth]{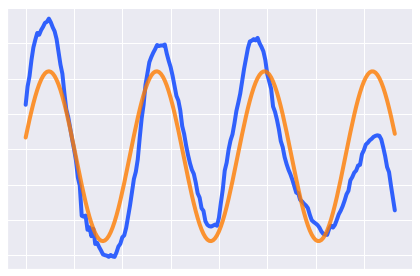}
    \includegraphics[width=0.11\linewidth]{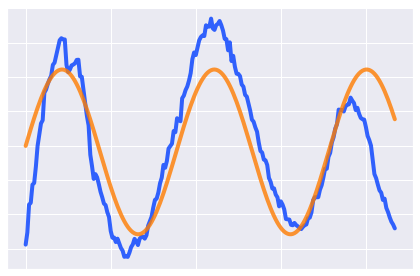}
    \includegraphics[width=0.11\linewidth]{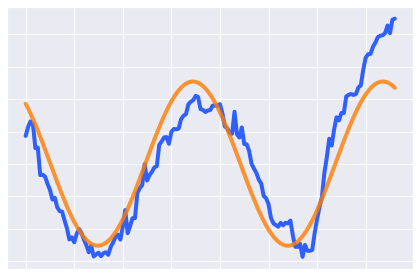}
    \includegraphics[width=0.11\linewidth]{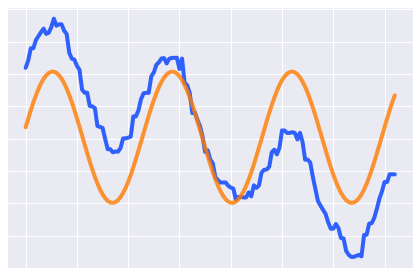}
    \includegraphics[width=0.11\linewidth]{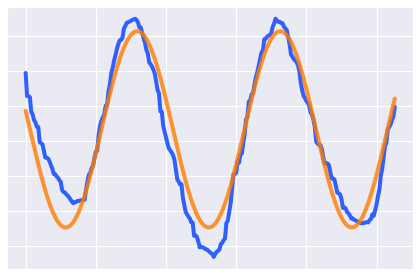}
    \caption{\textbf{Top:} Green dots correspond to keypoints identified by \name{} as relevant for determining the breathing rate. The red dots are discarded keypoints. Note that some the videos contain distractors that move in the scene (rollouts of scenes are shown in Fig.~\ref{fig:breathing_rollouts_appendix} in Appendix). \name{} with RANSAC is able to automatically discover regions of interest. \textbf{Bottom:} Timeseries (blue) and sinusoidal fit (orange) of one keypoint in the ROI for each of the scenes (same position in the $2\times4$ grid)}
    \label{fig:breathing_results}
\end{figure}

Fig.~\ref{fig:breathing_results} (top) shows the keypoints discovered for the 8 videos, with Fig.~\ref{fig:breathing_results} (bottom) showing the timeseries and its sinusoidal fit for one of the keypoints in the ROI. The model correctly identifies keypoints corresponding to the chest area, while ignoring distractor and lower-body keypoints. Comparing the respiratory periods identified with \name{} with the annotated values results in an MSE of 0.016 (in seconds/breath). In contrast, a baseline that uses the mean of the true rates for all videos obtains an MSE of 0.085. These results demonstrate the accuracy of \name{} for physical parameter estimation from an unknown region of interest, using only the knowledge that the motion of interest is sinusoidal as supervision.

\vspace{-4mm}
\section{Conclusion and future work}

This paper has introduced \name{}, a 3-stage method for dynamics-constrained keypoint discovery and system identification, which alternates between maximum likelihood extrinsic camera calibration and maximum likelihood physical parameter estimation for motion tracks detected in video. We enhance the stability of this optimization through the inclusion of a curriculum-based optimisation strategy, alongside a maximum entropy selection criterion for keypoint identification. Future avenues of work include extensions to multiple interacting objects, rigid or fluid body dynamics from video, and incorporation with a neural network for material and volume inference from vision.

{
\bibliographystyle{iccv/ieee_fullname}
\bibliography{iccv/references, iccv/additional_references}
}

\newpage

\appendix
\section{Discussion: Challenges and limitations}
\label{sec:limitations}
\keypoint{Scale unidentifiability} Due to the projection operation $\kk_t(\ttheta, \RR, \ttt) = [\Tilde{\pp}_{x,t}/\Tilde{\pp}_{z,t}, \Tilde{\pp}_{y,t}/\Tilde{\pp}_{z,t}]$, the 3D trajectory $\pp_{1:T}$ can only be determined up to a scale parameter. For this reason, we evaluate the correlation between the true- and learned parameters, not the error. This is also the metric used by GALILEO and Physics101 when doing physical parameter estimation from visual trajectories. Although scale unidentifiability leads to the existence of infinitely many solutions for $\ttheta$ and $\ttt$, and therefore instability with joint optimization, our use of alternate optimization steps guarantees that the algorithm converges to a single solution, as $\ttheta$ and $\ttt$ are optimized conditioned on one another, not jointly (this is akin to the Expectation-Maximization algorithm, where a marginal distribution is maximized via alternate optimization of conditionals).

\keypoint{Broken trajectories and occlusions:}
In settings where classical keypoint detectors are unreliable, one can either use state-of-the-art pretrained keypoint networks, like SuperPoint \cite{DeTone2018SuperPoint:Description} and LF-Net \cite{Ono2018LF-Net:Images}, or pretrain an unsupervised keypoint discovery network \cite{Jakab2018UnsupervisedGeneration,Minderer2019UnsupervisedVideos,Kulkarni2019UnsupervisedControl}. However, we found show that a simple grid keypoint detector yielded more reliable tracks than using classic (SIFT, ORB, FAST), or modern (SuperPoint, LF-Net) keypoint detectors.

In settings where standard optical flow computation is unreliable, more recent models (eg. FlowNet \cite{Fischer2015FlowNet:Networks}) could be used to provide flow estimates to the KLT tracker. More recent improvements to the KLT tracker (eg. CoMaL \cite{Ramakrishnan2016CoMaLBoundaries}) could also be used. The multi-stage, modular nature of our pipeline allows for the easy replacement of individual components, although we found standard optical flow computation to work very well in practice.

\keypoint{Camera roll set to zero:}
We found that setting the camera roll angle to zero greatly stabilized the optimization procedure. While this might be perceived as too strong of a constraint on the model, in the vast majority of real settings the camera has zero roll (i.e. it's rare for the camera to rotate around its projection axis). Therefore, imposing this constraint does not reduce the applicability of our method in the vast majority of cases, while providing improved results. Naturally, allowing roll optimization would make the model more general, but this is left as future work.


\newpage

\section{V-SysId pseudocode}
\label{sec:pseudocode}
\begin{algorithm}[H]
\caption{V-SysId}
\label{algo}
\begin{algorithmic}
\REQUIRE Video $V$ of length $T$
\REQUIRE Equation of motion $f$ of the object of interest
\REQUIRE KeypointTrackExtractor \; \# function that outputs a set of keypoint track proposals
\ENSURE Trajectory, physical parameters, and camera pose of the object of interest

\texttt{\\}

\# Get $N$ keypoint track proposals
\STATE Tracks $\leftarrow$ KeypointTrackExtractor($V$) 

\texttt{\\}

\# Fit physical parameters and camera pose to trajectory
\STATE $\text{SelectionCriterion} \leftarrow [\,]$
\STATE $\text{Params} \leftarrow [\,]$
\FOR{$n \in \{1...N\}$}

\STATE $\Tilde{\kk}_{1:T} \leftarrow \text{Tracks}[n]$
\STATE Initialize $\alpha \leftarrow 0$, $\beta \leftarrow 0$, $\ttt \leftarrow [0,0,0]$, $\vvv_0 \leftarrow [0,0,0]$;
\STATE Initialize $\pp_0$ as the projection of $\Tilde{\kk}_0$ onto the $z=5$ plane in world coordinates;
\STATE Initialize $\bm{\eta}$ to some sensible initial values (setting dependent);

\FOR{$t \in \{1...T\}$}
\STATE $\ttheta \leftarrow \argmax_{\ttheta} \; p(\Tilde{\kk}_{1:t}|\ttheta, \RR, \ttt)$
\STATE $\RR, \ttt \leftarrow \argmax_{\RR, \ttt} \; p(\Tilde{\kk}_{1:t}|\ttheta, \RR, \ttt)$
\ENDFOR

\STATE Append $\{\ttheta, \RR, \ttt\}$ to Params
\STATE Append the scalar $p(\Tilde{\kk}_{1:t}|\ttheta, \RR, \ttt) + H(\Tilde{\kk}_{1:T})$ to SelectionCriterion
\ENDFOR

\texttt{\\}

\# Trajectory selection
\STATE $n^* \leftarrow \argmax \text{SelectionCriterion}$

\STATE $\Tilde{\kk}^* = \text{Tracks}[n^*]$
\STATE $\ttheta^*, \RR^*, \ttt^* = \text{Params}[n^*]$

\RETURN $\Tilde{\kk}^*, \ttheta^*, \RR^*, \ttt^*$
\end{algorithmic}
\end{algorithm}

\subsection{Evaluating future trajectory prediction} 

\begin{figure}[th]
    \centering
    \includegraphics[width=0.43\linewidth]{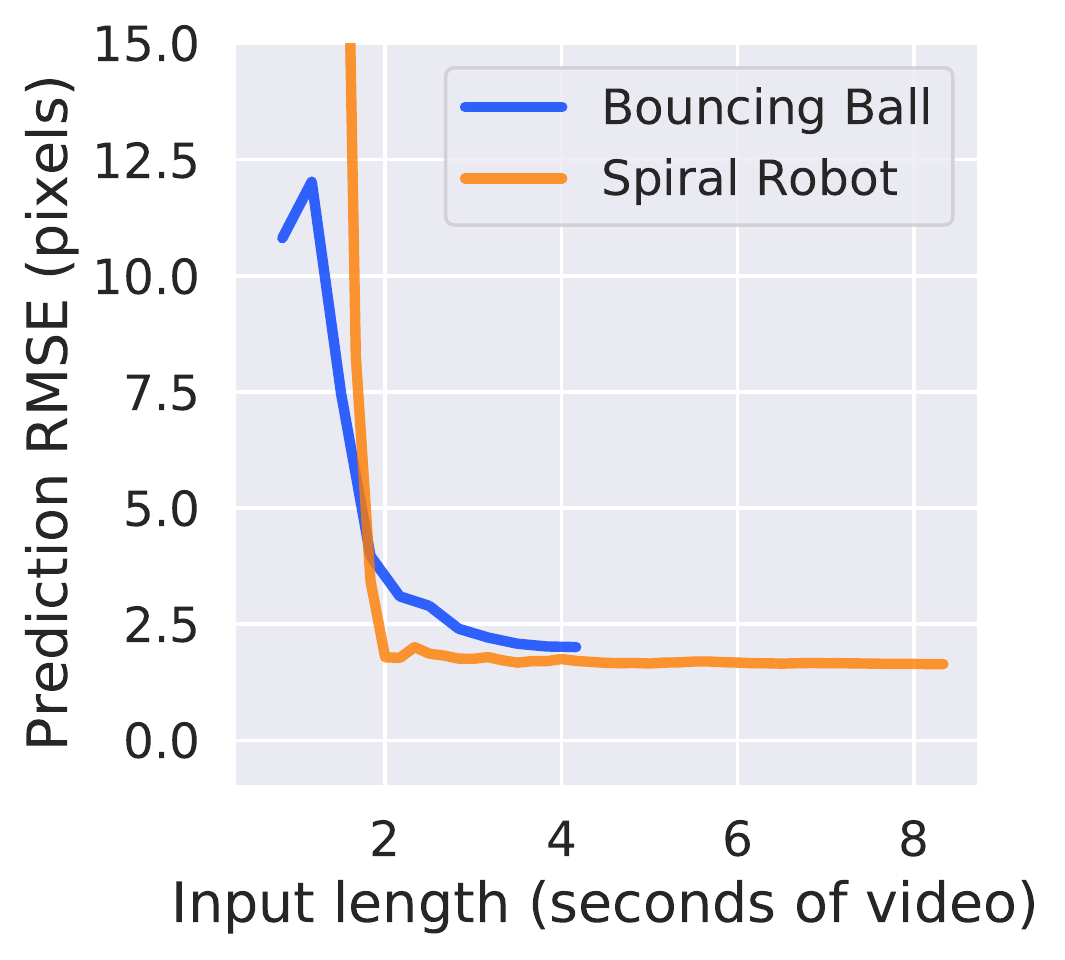}
    \caption{\textbf{Left:} Future trajectory prediction error under estimated parameters as a function of input length.}
    \label{fig:prediction}
\end{figure}

We now evaluate the ability of the optimization process of \name{} to perform accurate tracking and prediction given a sequence of correct keypoints. At each optimization iteration, we rollout the trajectory under the current parameters (i.e. those learned with the fraction of the sequence up to that iteration), and measure the prediction error relative to the observed keypoint track. This is done in hindsight only for the trajectory chosen by \name{}, although it could equally be done for a keypoint sequence inferred by a test-time inference neural network. The results for the bouncing ball and spiral robot are shown in Fig.~\ref{fig:prediction}. The curves show that the optimization process quickly converges to correct system identification, leading to correct trajectory prediction after only 2 seconds of input. 

\section{Comparison of keypoint detectors}
\label{sec:keypoint_ablation_appendix}
\begin{figure}[h]
    \centering
    \includegraphics[width=0.22\textwidth]{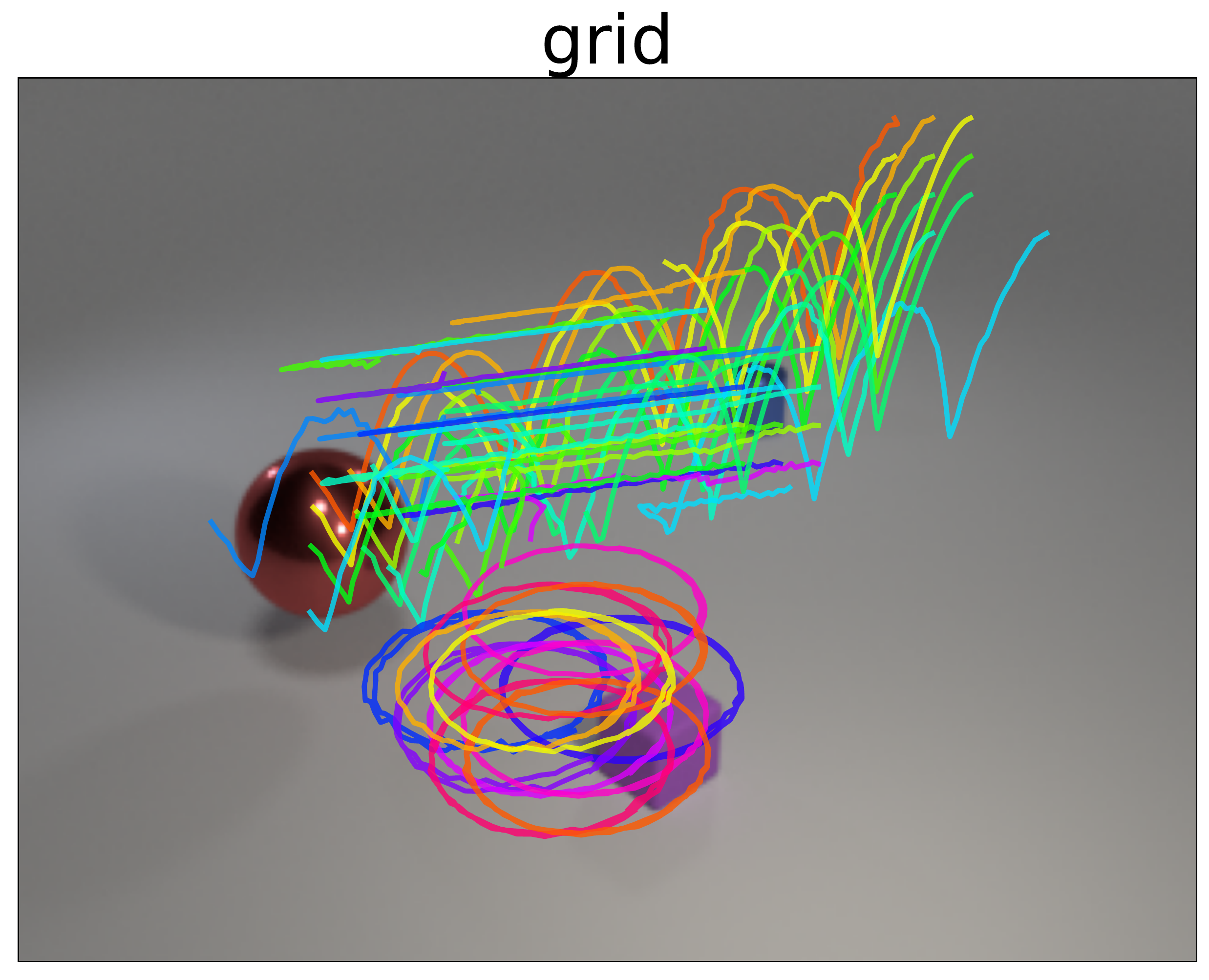}
    \includegraphics[width=0.22\textwidth]{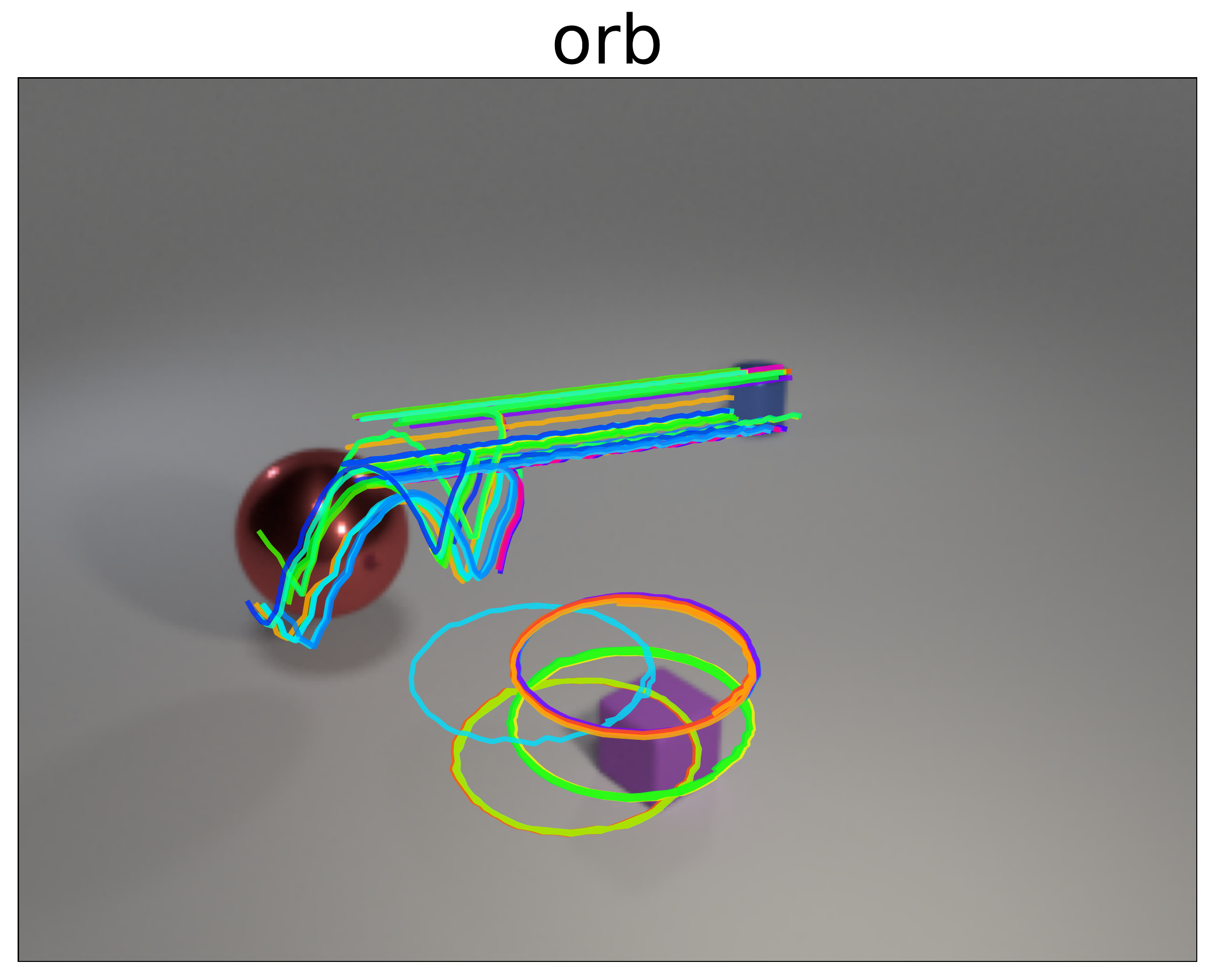}
    
    \hspace{0.05mm}
    \includegraphics[width=0.22\textwidth]{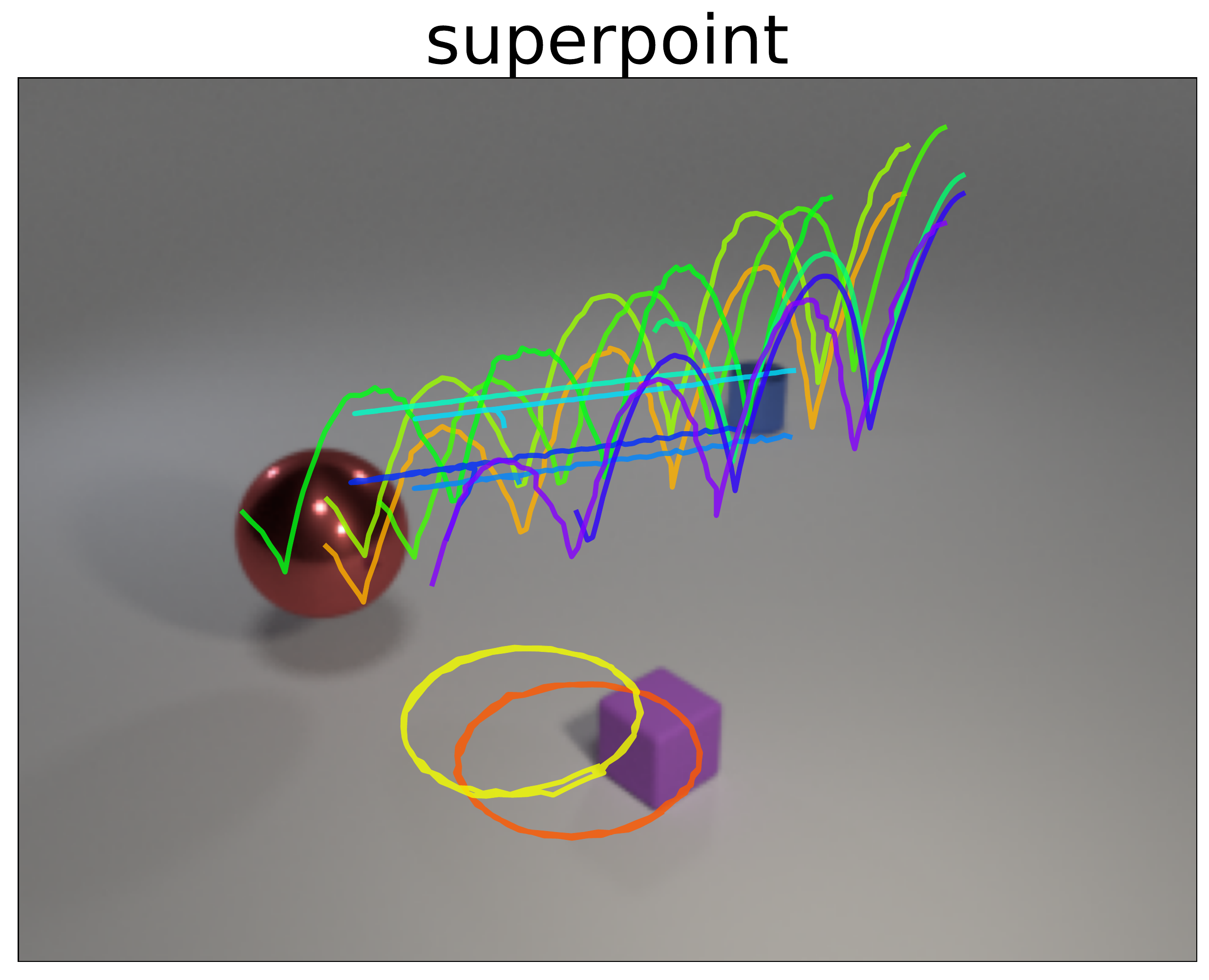}
    \includegraphics[width=0.22\textwidth]{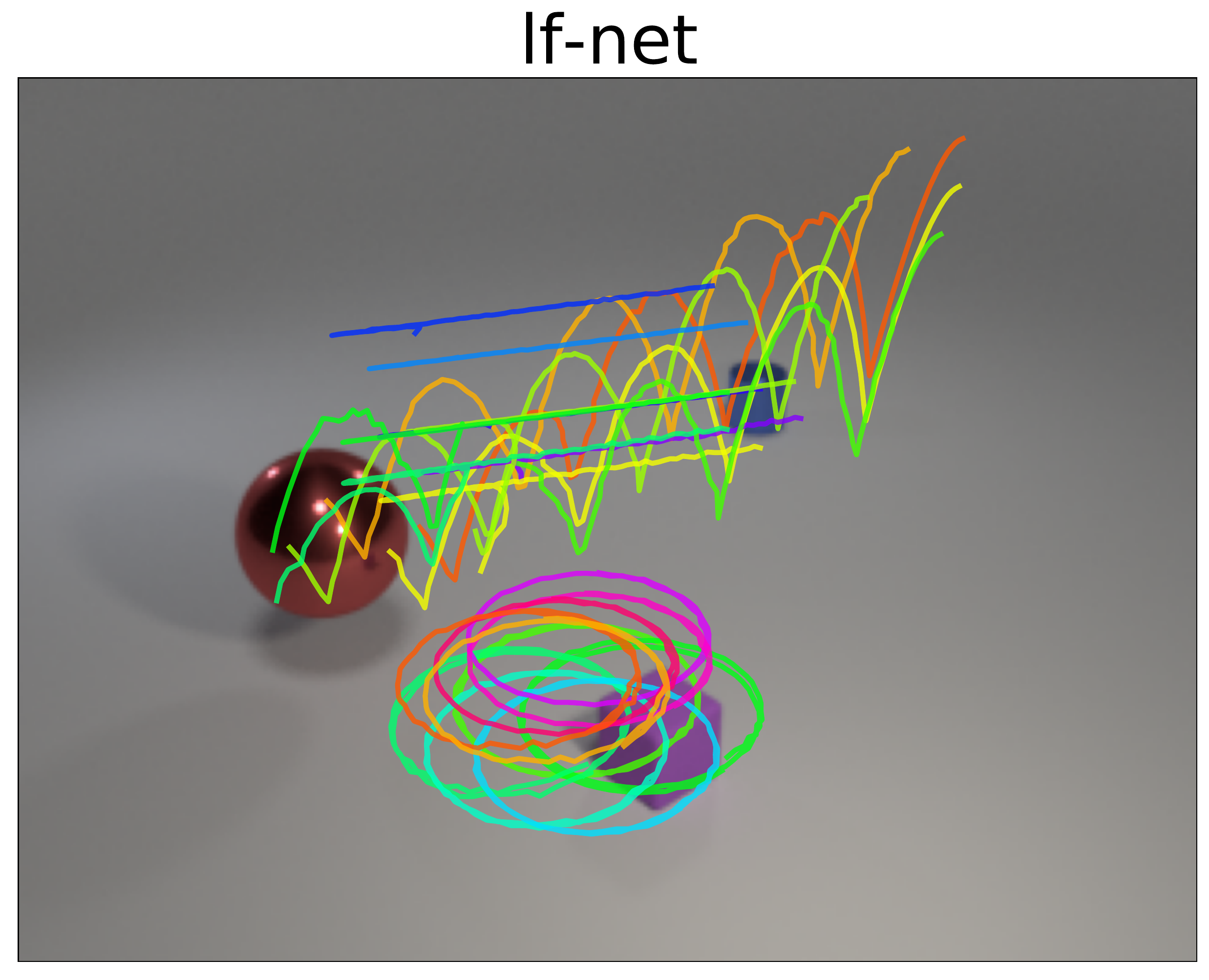}
    \caption{More visualization of the discovered object in various bouncing ball scenes.}
    \label{fig:keypoint_methods}
\end{figure}

Here we provide a visual comparison of the trajectory proposals obtained using grid, ORB, LF-Net and SuperPoint keypoint extractors, in conjunction with a KLT tracker. Fig.~\ref{fig:keypoint_methods} shows this comparison for the bouncing ball dataset (after filtering for short and static tracks). It can be seen that despite its simplicity, the grid extractor performs just as well as the more modern keypoint detectors, while running over an order of magnitude faster.


\section{Further visualizations}
See next page.

\begin{figure*}
  \centering
    \begin{minipage}{0.28\textwidth}
        \centering
        \small{Bouncing ball}
        \\[1mm]
        \includegraphics[width=\textwidth]{figures/blender_bb_keypoints.png}
    \end{minipage}%
    \hspace{2mm}
    \begin{minipage}{0.60\textwidth}
        \centering
        \includegraphics[width=\textwidth]{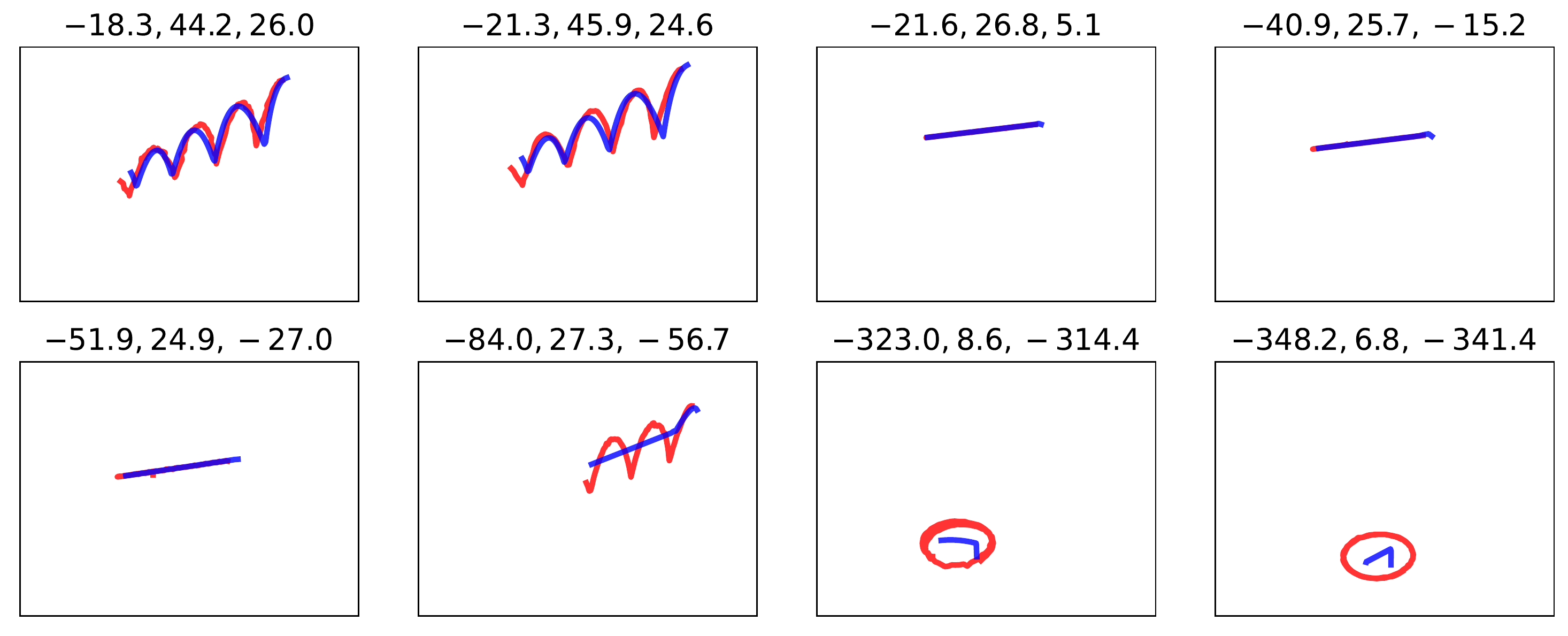}
    \end{minipage}
    
    \vspace{2mm}
    \begin{minipage}{0.28\textwidth}
        \centering
        \small{Robot arm end-effector}
        \\[1mm]
        \includegraphics[width=\textwidth]{figures/spiral_keypoints.png}
    \end{minipage}%
    \hspace{2mm}
    \begin{minipage}{0.60\textwidth}
        \centering
        \begin{overpic}[width=\linewidth]{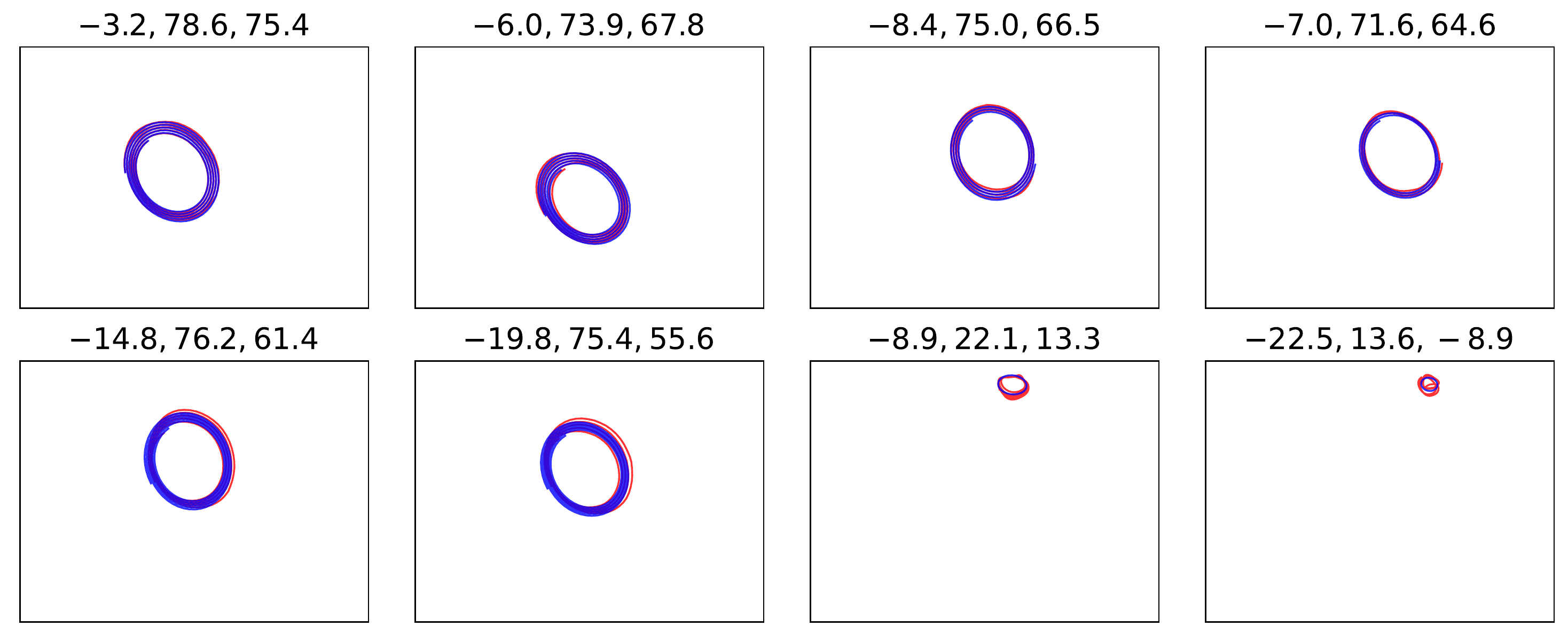}
         \put (15,24) {\small{\annot{Best}}}
         \put (60,5) {\scriptsize{\annot{Distractor}}}
         \put (85,5) {\scriptsize{\annot{Distractor}}}
         \put (15,67) {\small{\annot{Best}}}
         \put (60,67) {\scriptsize{\annot{Distractor}}}
         \put (85,67) {\scriptsize{\annot{Distractor}}}
         \put (11,47) {\scriptsize{\annot{Distractor}}}
         \put (60,54) {\scriptsize{\annot{Distractor}}}
         \put (85,54) {\scriptsize{\annot{Distractor}}}
        \end{overpic}
    \end{minipage}
    \vspace{2mm}
    \caption{\textbf{Left:} Keypoint tracks propsed by a grid keypoint detector + KLT tracker (short or static tracks not shown here for improved visualization). \textbf{Right:} Subset of the extracted keypoint tracks (red) and projected fitted trajectories (blue), with the corresponding projection loglikelihood, entropy, and their sum, over each plot.}
    \label{fig:more_trajs_and_stats}
\end{figure*}

\clearpage

\begin{figure*}
    \centering
    \includegraphics[width=0.96\textwidth]{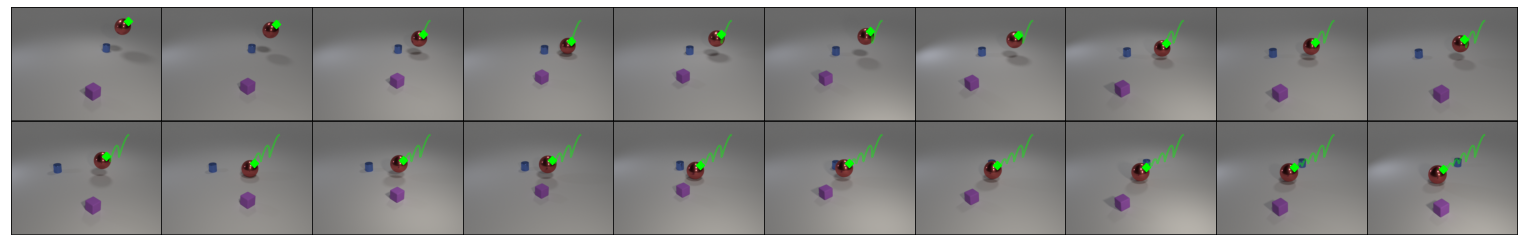}
    \includegraphics[width=0.96\textwidth]{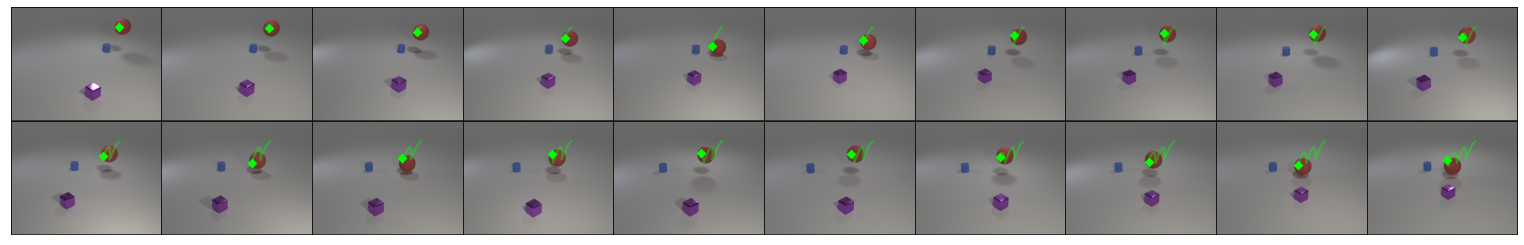}
    \includegraphics[width=0.96\textwidth]{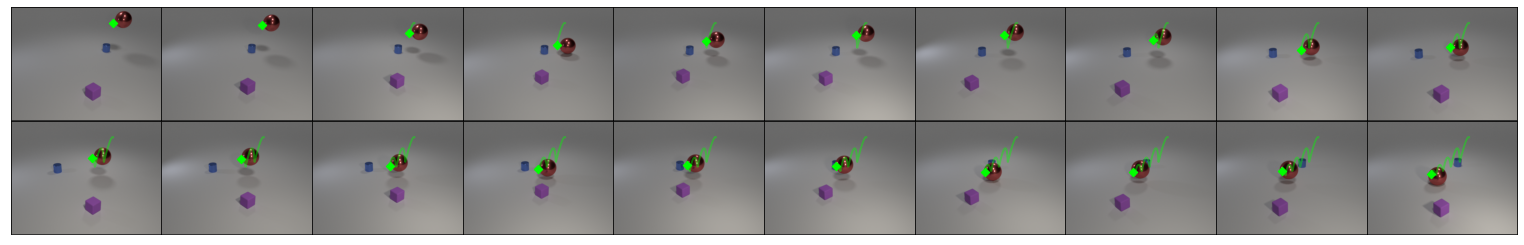}
    \includegraphics[width=0.96\textwidth]{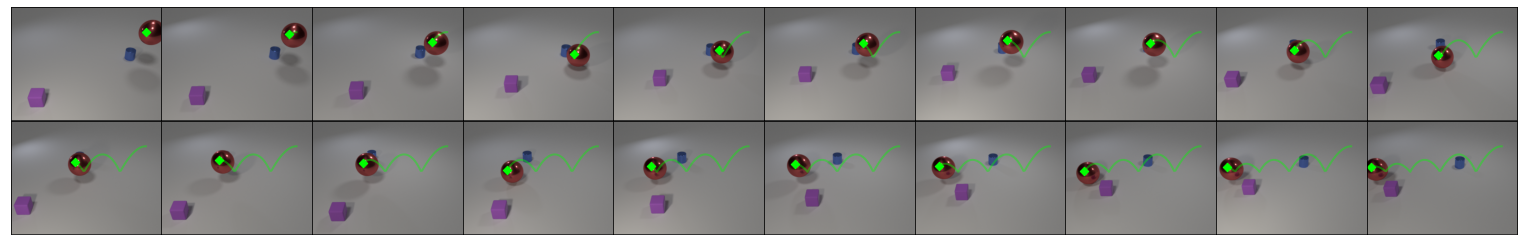}
    \hspace{0.05mm}
    \includegraphics[width=0.96\textwidth]{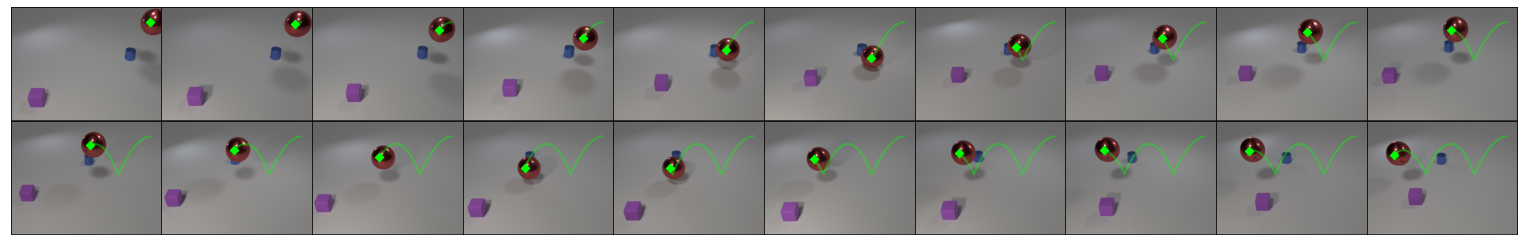}
    \caption{Comparison of various keypoints extractor and trackers on a bouncing ball scene.}
    \label{fig:bb_rollouts_appendix}
\end{figure*}

\begin{figure*}
    \centering
    \includegraphics[width=0.96\textwidth]{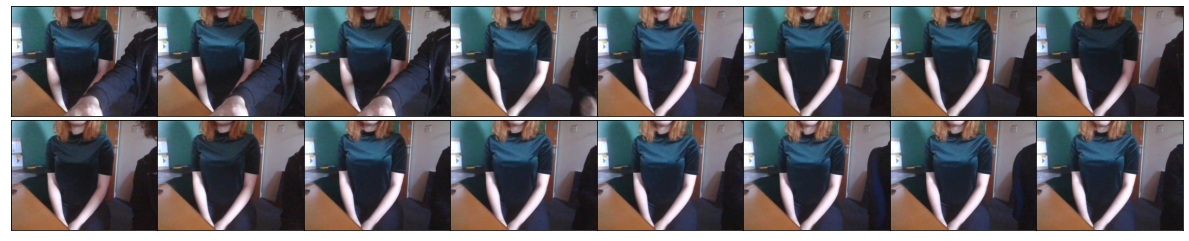}
    \hspace{0.05mm}
    \includegraphics[width=0.96\textwidth]{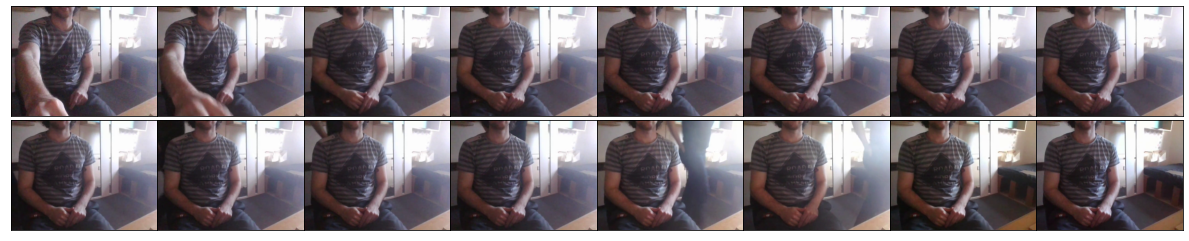}
    \caption{Frames of breathing scenes containing distractors.}
    \label{fig:breathing_rollouts_appendix}
\end{figure*}

\end{document}